\documentclass[10pt,journal,compsoc]{IEEEtran}

\usepackage[T1]{fontenc}
\usepackage{amsmath,amsfonts,amssymb}
\usepackage{graphicx}
\usepackage{booktabs}
\usepackage{cite}
\usepackage{algorithm}
\usepackage{algpseudocode}
\usepackage{listings}
\usepackage{xcolor}
\usepackage{caption}
\usepackage{hyperref}

\begin{document}
\title{A Formal Hierarchical Architecture for Agentic Orchestration with Stack-Based Execution and Lazy Discovery}

\author{Prashant~Devadiga,
Abhishek,
Adithya~Mishra,
Alok~Singh,
Amisha~Sinha,
Asit~Desai,
Gaurang~Dahad,
Harshit~Bhushan,
Mandati~Pramod~Reddy,
Prakhar~Gupta,
Rupesh~Patil,
and Siddhi~Behere%
\IEEEcompsocitemizethanks{%
\IEEEcompsocthanksitem Leadership: Prashant Devadiga.
\IEEEcompsocthanksitem Agentic Architecture: Abhishek, Alok Singh, Asit Desai, Gaurang Dahad, Harshit Bhushan, Prakhar Gupta.
\IEEEcompsocthanksitem Engineering: Adithya Mishra, Amisha Sinha, Mandati Pramod Reddy, Rupesh Patil, Siddhi Behere.
\IEEEcompsocthanksitem Evaluation: Abhishek, Gaurang Dahad, Harshit Bhushan.}%
}

\IEEEtitleabstractindextext{%
\begin{abstract}
\normalfont\normalsize
The rapid expansion of capabilities in Large Language Model (LLM) agents has exposed a critical architectural bottleneck: when agents are given access to a flat, monolithic registry of tools, the model must evaluate hundreds or thousands of options simultaneously. This leads to decision-space explosion, context window saturation, and degraded routing accuracy. To address these limitations, this paper presents a hierarchical, skill-based architecture for agentic orchestration. Capabilities are organized as a rooted tree where internal nodes make routing decisions and leaf nodes execute deterministic tasks. The runtime enforces a single-step execution loop governed by a Last-In-First-Out (LIFO) stack, giving the agent a form of memory akin to a Pushdown Automaton, therefore enabling it to track nested execution contexts and resume deterministically from any depth. Capability discovery follows a manifest-driven, lazy-loading protocol: only the immediate children of the active node are loaded, so memory and prompt costs scale with the explored path rather than the global registry. By replacing global memory with localized stack frames, the architecture prevents outputs from one execution branch from leaking into another, establishing the isolation guarantees required for deployment in regulated enterprise environments. We also discuss UPI Help, an AI-powered digital payments support product, as a motivating production deployment context. We provide a mathematical formalization of the orchestration state, detailed algorithmic analysis of the execution loop, and controlled benchmarks comparing flat and hierarchical routing under increasing tool catalogs, multi-step workflow pressure, and visible schema-token exposure per LLM call.
\end{abstract}}

\maketitle
\IEEEdisplaynontitleabstractindextext

\section{Introduction}

The integration of Large Language Models (LLMs) with external, deterministic capabilities (such as Application Programming Interfaces (APIs), database queries, and scripts) has enabled autonomous, goal-directed agents \cite{schick2023toolformer}. While early paradigms showed that interleaving reasoning with tool execution improves grounding and task completion, scaling these systems to enterprise-grade workflows exposes foundational architectural limitations \cite{liu2023agentbench}.

The primary limitation of contemporary LLM agents is the assumption of a flat capability space. Conventional agents initialize with a registry containing all accessible tools \cite{qin2023toolllm}. At a scale encompassing hundreds or thousands of distinct capabilities, this induces \textit{decision-space explosion}: the model must parse and choose among a large set of mostly irrelevant tools, increasing latency and degrading semantic routing accuracy \cite{du2024anytool}. Benchmarks confirm that large, non-hierarchical API arrays containing overlapping syntactic conventions increase reasoning failures and hallucinated tool invocations \cite{xu2024toolbench}.

To distribute cognitive load, many systems adopt \textit{multi-agent orchestration} frameworks \cite{wu2023autogen}. By deploying multiple specialized LLM agents that communicate via natural language, systems can theoretically tackle highly complex objectives \cite{hong2023metagpt}. However, these often rely on unstructured conversational message-passing for orchestration, which obscures execution state and makes routing and recovery difficult at scale, leading to unpredictable agent loops and network-wide context saturation \cite{qian2023chatdev}.

In addition, standard autonomous agents lack deterministic process control primitives (explicit call stacks and isolated memory frames) commonly used in operating systems to support nested execution and robust recovery \cite{packer2023memgpt}. In production, minor network disruptions or unexpected sub-tool outputs can derail long-running agentic loops, and untrusted outputs can pollute the global reasoning context, which is incompatible with high-trust deployments \cite{liu2023agentbench}.

The design presented in this paper is motivated by production requirements observed in UPI Help, an AI-powered digital payments support product for UPI transaction queries, complaint resolution, and mandate management \cite{npciupihelp}. In such settings, agentic orchestration must not only route accurately, but also preserve auditability, enforce capability boundaries, and support controlled execution of sensitive payment-support flows.

To address these interconnected bottlenecks across both flat-tool augmented models and unstructured multi-agent networks, we present a \textit{Hierarchical Skill-Based Architecture}. To establish academic rigor, we explicitly distinguish a ``Skill'' from a standard ``Tool'': a ``Tool'' is an isolated API endpoint exposed directly to the reasoning model, while a ``Skill'' is a contextually-bound node within a strict structural hierarchy. In this framework, capabilities are organized as an arbitrary-depth tree in which each node is either a \textit{decision-making skill} (responsible for localized orchestration and delegation) or an \textit{executable skill} (providing deterministic leaf-level execution).

The runtime mechanism is decoupled from capability definitions and executes a single-step loop governed by a LIFO stack. This stack discipline gives the agent a context-free memory structure (akin to a Pushdown Automaton), supporting nested execution and deterministic return semantics \cite{sipser2012introduction}. To avoid prompt bloat, capability discovery is manifest-driven and lazy: a node only loads the operational metadata of its immediate children when invoked. As a result, cognitive and computational load scales with the explored path, not the global capability set. Finally, capabilities are added declaratively via localized manifests, enabling horizontal (new siblings) and vertical (deeper sub-orchestrators) scaling without modifying the core routing loop.

\subsection{Paper Organization}
We will discuss the overall structure and details of our work as follows: Section 2 reviews related literature and situates our work within the broader context of multi-agent architectures and formal automata theory. Section 3 formulates the problem mathematically. Section 4 formalizes the system model and state representation. Section 5 presents the execution semantics and stack-based control loop. Section 6 describes capability discovery and lazy loading via localized manifests. Section 7 analyzes architectural properties and security guarantees. Section 8 presents UPI Help as a production deployment context. Section 9 reports experimental evaluation against flat baselines. Section 10 provides complexity and performance analysis. Section 11 discusses system constraints and limitations, Section 12 concludes the paper, and Section 13 lists contributors.

\section{Related Work}

We situate our architecture at the intersection of tool integration, multi-agent orchestration, automata-based formal verification, and capability discovery.

\subsection{Context and Scope: Tool Integration and Modularity}
The paradigm of equipping language models with external tools was heavily popularized by frameworks such as Toolformer \cite{schick2023toolformer}, which utilized self-supervised fine-tuning to train models to emit API calls during text generation. While foundational, these systems operated entirely within flat integration schemas, suffering from rapid context saturation when exposed to vast operational libraries \cite{liang2023taskmatrix}. To mitigate this, subsequent embodied architectures, most notably Voyager \cite{wang2023voyager}, introduced the concept of an ever-growing, composable skill library, proving that modularizing task-specific code into retrievable discrete units drastically mitigates catastrophic forgetting. 

As the scale of required APIs surged to the millions, systems like TaskMatrix.AI \cite{liang2023taskmatrix} demonstrated that static prompt injection of tools was impossible, necessitating external metadata retrieval mechanisms. Consequently, hierarchical orchestration emerged. Frameworks such as AnyTool \cite{du2024anytool} developed self-reflective, hierarchical agents capable of navigating large-scale API ecosystems without incurring prompt bloat. Concurrently, compiler-inspired frameworks like LLMCompiler \cite{kim2024llmcompiler} proved the efficacy of permanently decoupling the high-level task planner from the deterministic execution units, achieving massive reductions in latency. Our implemented architecture builds upon these concepts by enforcing a strict bifurcation between decision-making orchestration nodes and executable deterministic leaves.

\subsection{Multi-Agent Orchestration Architectures}
Multi-agent orchestration frameworks, such as AutoGen \cite{wu2023autogen} and MetaGPT \cite{hong2023metagpt}, distribute work across specialized agents, where orchestration is typically implemented as conversational message-passing. Systems such as ChatDev \cite{qian2023chatdev} show that role specialization (e.g., planner, coder, reviewer) can improve output quality, but this makes it difficult to recover deterministically, audit execution paths, or prevent long-horizon conversational drift. Our architecture addresses this by treating decision-making skills as nodes in a governed hierarchical call stack with explicit return semantics, rather than as conversational peers.

\subsection{Theoretical Automata and Agentic Memory}
The stability of long-horizon agents is shaped by their memory model. Prior work has related common agentic frameworks to the Chomsky hierarchy: agents that rely solely on a flat, rolling context resemble Regular Agents (Finite State Machines), while agents with explicit LIFO memory can represent context-free control patterns (Probabilistic Pushdown Automata) \cite{sipser2012introduction}. Systems such as MemGPT \cite{packer2023memgpt} further argue for operating-system-inspired context management, such as paging and interrupts, which enable deterministic and lossless context switching. Our work adopts this lens to formalize and implement stack-based control with localized memory frames.

\subsection{Capability Discovery and Governance}
In decentralized ecosystems, hard-coding capabilities into an agent's core loop limits scalability and operational agility \cite{zhang2023agentverse}. The industry has rapidly shifted toward dynamic capability discovery using standardized declarative manifests. The Model Context Protocol (MCP) by Anthropic \cite{anthropic2025mcp} formalizes this by introducing JSON-RPC communication rules where agents discover contextual tools dynamically. These protocols mirror classical database optimization research, specifically the workload-driven, lazy discovery of data dependencies \cite{graefe1993query}.

\subsection{Our Contributions}
Taken together, existing approaches still commonly expose large flat tool spaces, rely on informal conversational routing, or lack explicit stack-like control and localized memory needed for deep, auditable workflows \cite{xu2024toolbench, liu2023agentbench}. To address this gap, we present a hierarchical orchestration architecture with the following contributions:
\begin{enumerate}
    \item \textbf{Formal System Model:} We propose a topological capability tree that permanently decouples probabilistic decision-making orchestrators from deterministic execution leaves.
    \item \textbf{Stack-Based Execution Semantics:} We implement a single-step, context-free execution loop governed by a LIFO stack, mathematically elevating the agentic workflow to a Pushdown Automaton capable of lossless resumption.
    \item \textbf{Manifest-Driven Lazy Discovery:} We establish a localized discovery protocol that restricts capability loading to the active execution path, mitigating prompt bloat.
    \item \textbf{Execution Isolation and Enterprise Governance:} We demonstrate that hierarchical capability fencing structurally prevents prompt injection persistence, making the architecture viable for high-trust domains.
    \item \textbf{Declarative, Routing-Agnostic Scaling:} We provide a framework where adding new functionality (scaling either horizontally with new sibling tools or vertically with deeper sub-orchestrators) is achieved via purely declarative means using JSON manifests, requiring zero modifications to the core orchestration logic.
    \item \textbf{Production Deployment Context:} We discuss UPI Help as a motivating production setting for applying these ideas in digital payments support.
    \item \textbf{Empirical Evaluation:} We benchmark flat versus hierarchical routing on prompt scalability, visible schema exposure, workflow pressure, and production-adjacent UPI flows, showing that hierarchy bounds per-step context while accuracy depends on tree design.
\end{enumerate}

\section{Problem Formulation}

We formalize two bottlenecks in flat-registry agent architectures and derive design objectives for a hierarchical solution.

\subsection{Contextual Saturation}
Let $V$ represent the global set of all available deterministic capabilities (tools or APIs) within an enterprise system, where the total scale is defined as $|V| = N$. In a standard flat-agent architecture, denoted as $\mathcal{A}_{\text{flat}}$, the agent acts as a single centralized orchestrator node connected directly to all $N$ leaves. 

To enable autonomous semantic routing, the inference engine must embed the metadata and functional schema of every tool into the active context window $\Omega$ at every reasoning step $t$. The token footprint of the prompt is thus defined as:
\begin{equation}
|\Omega_t|_{\text{flat}} = \sum_{v \in V} |\text{schema}(v)| + |\mathcal{H}_t|
\end{equation}
where $\mathcal{H}_t$ represents the accumulated dialogue history. As $N$ grows, the condition $|\Omega_t| \ge L_{\max}$ (the LLM's token limit) is eventually met. Even before strict token exhaustion, exposing the model to $N$ simultaneous capabilities triggers a \textit{decision-space explosion}, degrading routing accuracy and inducing hallucinated tool calls \cite{xu2024toolbench}.

\subsection{Linear State Fragility}
Standard autonomous agents treat execution state entirely as an unstructured, append-only dialogue array:
\begin{equation}
\mathcal{H} = \langle m_0, m_1, m_2, \dots, m_t \rangle
\end{equation}
Because this representation has no hierarchical depth indicators or localized memory partitions, it behaves like a Regular Agent (Finite State Machine). If a multi-step sub-task fails at step $t$, the flat context history provides no explicit mechanism to resume from the start of that sub-task.

\subsection{Design Objectives}
To resolve these interconnected bottlenecks, the objective of the proposed architecture is to replace $\mathcal{A}_{\text{flat}}$ with a graph-based framework $\mathcal{A}_{\text{hierarchical}}$ that satisfies two strict conditions:
\begin{enumerate}
    \item \textbf{Sub-Linear Prompt Scaling:} The context window $|\Omega_t|$ must scale entirely independent of the global registry size $N$.
    \item \textbf{Context-Free State Resumption:} The system must incorporate pushdown memory primitives (a call stack) to isolate localized task data and enable deterministic error recovery without contaminating the global dialogue history $\mathcal{H}$.
\end{enumerate}

\section{System Model and Topological Architecture}

The system relies on a structural abstraction that decouples the statefulness of the multi-agent workflow from the underlying reasoning engine. We formalize the architectural topology and the centralized state representation that support arbitrary-depth orchestration.

\subsection{The Capability Tree}
The architecture represents all systemic operations as a finite, rooted tree $\mathcal{T} = (V, E, r)$, where $V$ represents the set of all available skills, $E \subseteq V \times V$ represents the localized parent-child invocation relationships, and $r \in V$ is the root orchestrator skill. 

To enforce strict separation of concerns, the system mandates that every skill $v \in V$ must resolve to one of two mutually exclusive functional types:
\begin{enumerate}
    \item \textbf{Decision-Making Skill ($v_\delta$):} An internal orchestrator node where the set of children $children(v_\delta) \neq \emptyset$. These nodes do not interact with external systems directly; their sole responsibility is localized reasoning, planning, and delegating intents to their immediate child sub-skills.
    \item \textbf{Executable Skill ($v_\epsilon$):} A terminal leaf node implementing deterministic, functional logic (e.g., API requests, SQL database queries, computational scripts). These nodes execute pre-defined Python routines and possess no children ($children(v_\epsilon) = \emptyset$).
\end{enumerate}

The hierarchical depth $d(v)$ represents the edge distance from the root orchestrator $r$ to the current node $v$. 

\subsection{State Representation and Directed Graph Memory}
Standard LLM wrappers typically pass lists of dialogue dictionaries to the inference endpoint. To support arbitrary-depth nested workflows without losing operational context, our architecture maintains a centralized state and represents execution as a state graph. Rather than relying on procedural, hard-coded execution paths, the orchestration logic is defined by uniform transitions between execution states, with edges capturing conditional routing logic.

Instead of maintaining a single global memory scratchpad, the global state $\mathcal{S}$ persists across transitions and is constrained to routing and top-level interactions. The state is mathematically formulated as a tuple:
\begin{equation}
\mathcal{S} = \langle \mathcal{M}, \Gamma, p_{call} \rangle
\end{equation}

Where:
\begin{itemize}
    \item $\mathcal{M}$ represents the core conversation state, maintaining the ordered array of human-agent interaction histories, specific chat identifiers, and standardized user metadata.
    \item $\Gamma$ defines the \textbf{Skill Path Stack}, an ordered LIFO list of active operational frames, explicitly providing the pushdown memory required for recursive tool invocation across graph cycles.
    \item $p_{call}$ represents the \textbf{Pending Skill Call}, a pointer defining the identifier, input schema, and target destination of the next node queued for execution.
\end{itemize}

\subsection{The Skill Context Frame}
Every element within the LIFO stack $\Gamma$ is instantiated as a \texttt{SkillContext} object. This object acts as a localized execution snapshot, capturing the metadata of an active node and logically isolating memory at the branch level as control flows through the state graph. Let a context frame $\gamma_i \in \Gamma$ be defined as:
\begin{equation}
\gamma_i = \langle \mathit{id}, \rho, \mathit{parent}, \tau, o_{\mathit{self}}, \mathcal{O}_{\mathit{acc}} \rangle
\end{equation}

Here, $\mathit{id}$ is the operational skill name, and $\rho$ is the absolute file system path mapping the node to its directory. The $\mathit{parent}$ maintains a back-reference to the calling orchestrator. The parameter $\tau$ dictates the conditional edge logic that the primary execution loop applies to this frame.

Crucially, memory is bifurcated within the frame itself. $o_{\mathit{self}}$ represents the output produced by the skill upon its own completion. Conversely, $\mathcal{O}_{\mathit{acc}}$ introduces a \textbf{localized output accumulator}. As the state graph routes execution back to a parent node, the localized accumulator preserves the results of its immediate children. This graph-bound, frame-based memory reduces cross-branch contamination and controls data visibility during complex multi-step traversals.

\section{Execution Semantics and The Control Automaton}

The execution runtime fundamentally discards the traditional approach of embedding hard-coded conditional logic for every possible tool combination. Instead, it relies on a uniform, single-step execution loop constructed via a formal state machine graph. This loop identically evaluates state transitions regardless of hierarchical depth, governed entirely by the data resident in $\Gamma$ and $p_{call}$.

\subsection{The Stack-Pending Invariant}
To ensure predictable execution tracing and auditable state reconstruction, the runtime preserves a mathematical invariant throughout the execution lifecycle:

\textbf{Proposition 1 (Stack-Pending Invariant):} \textit{If the pending invocation pointer $p_{call}$ is populated with a valid target node $v$, and $v \neq r$ (where $r$ is the root), then the top frame of the stack $\gamma_{top} \in \Gamma$ strictly corresponds to the execution context of $parent(v)$.}

This invariant guarantees that whenever control is passed to a new skill, its parent's state is safely preserved on the stack, mimicking the execution context preservation required for verifiable context switching in programmatic function calls.

\subsection{Single-Step Orchestration Algorithm}
The core operational loop is encapsulated within the primary skill processing node. Upon entry, the node dynamically delegates execution based on the functional type parameter $\tau$ of the active skill. The control flow is formalized in Algorithm \ref{alg:execution_loop}.

\begin{algorithm}
\caption{Uniform Single-Step Execution Loop}\label{alg:execution_loop}
\begin{algorithmic}[1]
\Require Initialized State $\mathcal{S}$ with $\Gamma = \emptyset, p_{call} = r$
\Ensure A deterministic terminal output and updated global state $\mathcal{S}$
\While{$p_{call} \neq \text{None}$}
    \State Let $s \gets p_{call}$
    \State Clear $p_{call}$ in $\mathcal{S}$
    \State Let context $\gamma_s \gets \text{InitializeContext}(s)$
    
    \If{$\gamma_s.\tau = \text{Decision-Making}$}
        \State Push $\gamma_s$ onto $\mathcal{S}.\Gamma$
        \State $M_{child} \gets \text{LazyDiscoverManifest}(s)$
        \State $prompt \gets \text{CompilePrompt}(M_{child}, \gamma_s.\mathcal{O}_{\mathit{acc}})$
        \State $response \gets \text{LLM.Generate}(prompt)$
        
        \If{$response.\text{type} = \text{SKILL\_CALL}$}
            \State $\mathcal{S}.p_{call} \gets \text{BuildPath}(s, response.\text{target})$
        \ElsIf{$response.\text{type} = \text{OUTPUT}$}
            \If{$response.\text{AllTasksComplete} = \text{True}$}
                \State \Return $response.\text{output}$
            \Else
                \State $\gamma_{\mathit{parent}} \gets \text{Pop}(\mathcal{S}.\Gamma)$
                \State $\mathcal{S}.p_{call} \gets \gamma_{\mathit{parent}}$
                \State Append $response.\text{output}$ to $\gamma_{\mathit{parent}}.\mathcal{O}_{\mathit{acc}}$
            \EndIf
        \EndIf
        
    \ElsIf{$\gamma_s.\tau = \text{Executable}$}
        \State $result \gets \text{ExecuteDeterministicLogic}(s)$
        \State $\gamma_{\mathit{parent}} \gets \text{Peek}(\mathcal{S}.\Gamma)$
        \State Append $result$ to $\gamma_{\mathit{parent}}.\mathcal{O}_{\mathit{acc}}$
        \State $\mathcal{S}.p_{call} \gets \gamma_{\mathit{parent}}$
        \State $\text{Pop}(\mathcal{S}.\Gamma)$
    \EndIf
\EndWhile
\end{algorithmic}
\end{algorithm}

\subsection{Analysis of Flow Control Mechanisms}
Algorithm \ref{alg:execution_loop} is designed so that the LLM makes \emph{local decisions} (which child skill to invoke), while control-flow book-keeping (how to return, what state is visible, and when to terminate) is handled by the runtime. This division improves predictability during deep traversals and reduces the chance of navigation errors \cite{liu2023agentbench}.

\textbf{Automatic Execution Bubbling:} When an executable leaf completes its deterministic processing (Algorithm 1, Line 22), the runtime returns control to the parent by restoring the parent frame from the LIFO stack, rather than asking the LLM to infer the correct next hop. This makes the return path deterministic: the model cannot ``skip levels,'' branch into unrelated capabilities, or lose its place in a nested workflow. In practice, this is a key mechanism for auditability and for preventing spurious transitions in long-horizon executions.

\textbf{Localized Output Bubbling:} A defining aspect of this architecture is that the child results are appended only to the parent frame's localized accumulator $\mathcal{O}_{\mathit{acc}}$, not to a global scratchpad. As a result, each decision-making node conditions on exactly the outputs it explicitly requested from its immediate children. This enforces a clear data-visibility boundary between branches of the capability tree, limiting cross-branch interference and reducing the blast radius of malformed or adversarial tool outputs.

\textbf{Global Task Completion Routing:} The loop supports a completion flag (e.g., \texttt{AllTasksComplete}) so that once the global objective is resolved, execution can terminate without bubbling a final answer through every ancestor frame. This preserves the stack discipline for partial sub-task completion, while avoiding unnecessary additional LLM turns when a leaf (or intermediate) skill has already produced the final user-facing result. Conversely, if the global objective is not yet met, the runtime assumes only a localized sub-task is finished and correctly pops the stack to return control to the parent frame.

\section{Capability Discovery and Manifest Discipline}

A defining aspect of the architecture is the decoupling of the runtime engine from the specific programmatic capabilities it executes. Standard agents often use monolithic global registries, loading large sets of API schemas into the system prompt at initialization \cite{qin2023toolllm}. In contrast, our approach uses progressive disclosure powered by localized JSON manifests and a lazy-loading discipline.

\subsection{Localized Manifest Architecture}
Each decision-making skill directory carries a \textbf{localized child manifest}, conventionally \texttt{skills\_info.json}, that lists only its immediate, direct children (not the global registry). The file is a JSON object mapping skill keys to entries; lazy discovery parses only the manifest at the active path $\rho$ (each nested orchestrator or leaf level repeats the same pattern).

A manifest \emph{entry} is defined by:
\begin{equation}
\label{eq:manifest}
M = \{ \mathit{id}, \mathit{desc}, \tau, \Sigma_{in}, \Sigma_{out}, \mu \}
\end{equation}
mapping to \texttt{name}, \texttt{description}, \texttt{metadata.has\_sub\_skills} ($\tau$: decision-making vs.\ executable), \texttt{input\_schema}, \texttt{output\_schema}, and governance fields in \texttt{metadata} (e.g., \texttt{requires\_auth}, \texttt{is\_terminal}).

Listing~\ref{lst:manifest} excerpts our UPI Help deployment: the root orchestrator entry \texttt{mandate} (\texttt{has\_sub\_skills}: \texttt{true}, empty schemas) and a mandate-level executable leaf \texttt{mandate\_summary} (\texttt{has\_sub\_skills}: \texttt{false}, populated schemas). The leaf entry is taken from \texttt{skills/mandate/skills\_info.json}; both entries illustrate the same manifest schema.

\begin{lstlisting}
{
  "mandate": {
    "name": "mandate",
    "description": "Handles UPI mandate (AutoPay) operations...",
    "input_schema": {}, "output_schema": {},
    "metadata": { "has_sub_skills": true, "requires_auth": true }
  },
  "mandate_summary": {
    "name": "mandate_summary",
    "description": "Fetches and lists all UPI mandates.",
    "input_schema": {
      "type": "object",
      "properties": { "mobile_number_hash": {"type": "string"} }
    },
    "output_schema": { "type": "object" },
    "metadata": { "has_sub_skills": false, "mcp_tool": true }
  }
}
\end{lstlisting}
\captionof{lstlisting}{Excerpt of \texttt{skills\_info.json}.}
\label{lst:manifest}
\medskip

By defining capabilities within localized manifests, the architecture aligns with modern capability protocols such as the Model Context Protocol (MCP) \cite{anthropic2025mcp}. Capabilities are exposed only when contextually relevant, analogous to the use of standardized API specifications such as OpenAPI \cite{openapi2021}.

\subsection{Lazy Discovery Logic}
The discovery module is triggered only when the execution loop enters a new decision-making node. The discovery protocol executes a strict localized parse, as outlined in Algorithm \ref{alg:discovery}.

\begin{algorithm}
\caption{Lazy Capability Discovery}\label{alg:discovery}
\begin{algorithmic}[1]
\Require Current skill node $v$ with path $\rho$ and functional type $\tau$
\If{$v.\tau = \text{Executable}$}
    \State \Return $\emptyset$ \Comment{Executable nodes have no children}
\ElsIf{$v.\tau = \text{Decision-Making}$}
    \State $M_{raw} \gets \text{ParseJSON}(\rho + \text{'/skills\_info.json'})$
    \State $Children \gets \emptyset$
    \For{each $c \in M_{raw}$}
        \State \Comment{Construct standardized manifest for child per equation 5}
        \State $m_{\mathit{child}} \gets \{ \mathit{id}: c.\mathit{id}, \mathit{desc}: c.\mathit{desc}, \tau: c.\tau, \Sigma_{in}: c.\Sigma_{in}, \Sigma_{out}: c.\Sigma_{out}, \mu: c.\mu \}$
        \State Append $m_{\mathit{child}}$ to $Children$
    \EndFor
    \State \Return $Children$
\EndIf
\end{algorithmic}
\end{algorithm}

This lazy-loading approach is consistent with classical database methodologies in which query evaluation can benefit from workload-driven, on-demand discovery of data dependencies \cite{graefe1993query}. By ensuring that the memory footprint and the LLM context prompt scale with the explored execution branch, the system mitigates the context-saturation problem.

\section{Architectural Properties and Security Guarantees}

The stack-based architecture provides structural properties that are useful beyond basic multi-agent orchestration. By combining explicit memory boundaries with declarative metadata, the framework supports secure and extensible enterprise deployments.

\subsection{Routing-Agnostic Extensibility}
A primary implication of this architecture is the decoupling of routing logic from the functional capability set. Scaling vertically (adding deeper sub-orchestrators) or horizontally (adding sibling executable tools) can be achieved by adding localized directories and appending declarative JSON entries to the parent's manifest, without modifying the core orchestration loop. This separation of concerns supports decentralized capability development while reducing regressions in the central routing logic.

\subsection{Capability Fencing and Attack Surface Reduction}
This mechanism limits the \emph{available actions} at each step by restricting which capabilities are visible to the decision-making model. As established in Section 6, the lazy discovery protocol restricts the context window to the immediate children of the active orchestrator. Beyond computational efficiency, this structural isolation supports the principle of least privilege \cite{anthropic2025mcp} and reduces the attack surface for unintended tool invocation.

Let $V$ represent the global set of all system capabilities, where $|V| = N$, and let $\Omega_t$ denote the aggregate schema representation injected into the LLM context at inference step $t$. Because the active node only loads its immediate children, we have $|\Omega_t|_{\text{hierarchical}} \ll |\Omega_t|_{\text{flat}}$ in typical deployments. This bounds the decision space to the node's local out-degree and limits the ability of a compromised or hallucinating node to invoke unrelated capabilities outside its current scope.

\subsection{Execution Isolation and Prompt Injection Mitigation}
This mechanism limits \emph{information flow} across steps by scoping intermediate outputs to stack frames and controlling how results bubble upward.
The hierarchical isolation of nodes replicates the security architecture established in systems like MemGPT \cite{packer2023memgpt}, which draw heavily from operating-system process isolation. When an executable leaf node runs a script to fetch external data (e.g., retrieving an untrusted user email), the raw payload is processed entirely within that terminal node's isolated execution frame. 

Because the state graph's transition logic relies on schema contracts and type validation to bubble outputs into the parent's localized accumulator ($\mathcal{O}_{\mathit{acc}}$), untrusted strings can be constrained to leaf-level handling. Thus the orchestrator receives only the validated intent injected into its isolated memory frame. This compartmentalization reduces the likelihood of indirect prompt injection persisting across the global reasoning state \cite{packer2023memgpt}.

\section{Production Deployment Context: UPI Help}

Although the architecture is formalized generically, it is motivated by deployment requirements from UPI Help, an AI-powered support product for UPI-related transaction assistance, complaint resolution, and mandate management \cite{npciupihelp}. This section uses UPI Help as a production context to illustrate why hierarchical orchestration, lazy discovery, and localized execution state are useful in regulated payment-support environments.

\subsection{Agentic Architecture in UPI Help}
UPI Help can be viewed as a domain-specific agentic system in which user requests are first routed to broad payment-support domains and then delegated to narrower operational capabilities. For example, high-level user intents may involve transaction status checks, dispute or complaint workflows, or recurring-payment mandate operations. A monolithic agent would expose all of these capabilities at once, while the hierarchical architecture separates them into localized decision-making skills and deterministic executable skills. This allows the root orchestrator to select the relevant domain, after which only that domain's immediate sub-skills are exposed for subsequent routing.

\subsection{Observed Benefits over Monolithic Routing}
Controlled benchmarks in the UPI Help deployment context (Section~9) indicate that hierarchical orchestration can improve routing stability by reducing irrelevant tool exposure and preserving local task context. On mandate and payment query families where the skill tree is fully populated, the hierarchical implementation achieves 100\% route-level success with lower per-query LLM latency than a flat terminal baseline. At large synthetic catalog sizes ($N \geq 256$ tools with 800-character schemas), flat routing becomes context-infeasible while hierarchical routing remains viable by bounding visible schema tokens per call.

Qualitatively, transaction-status queries benefit from domain-first routing: the root selects the payment domain before exposing transaction-specific executables, rather than presenting mandate and complaint tools simultaneously. Mandate flows similarly narrow the decision space to mandate-specific skills, preserving intermediate outputs within localized frames. Multi-step resolutions (e.g., fetch metadata, check status, escalate complaint) rely on stack-based return semantics so child results bubble to the parent orchestrator without reconstructing execution paths from flat dialogue history.

\subsection{Governance and Security Controls}
The production setting also motivates governance controls that are difficult to enforce cleanly in flat registries. Capability fencing limits which tools are visible at each step, reducing unintended cross-domain invocation. Localized execution frames limit how intermediate outputs and untrusted tool results propagate through the system. Sensitive flows, such as mandate modification or complaint escalation, can be guarded through schema validation, confirmation requirements, audit logging, and manifest-level governance metadata. These controls align the architecture with the operational requirements of high-trust payment-support systems while keeping the formal model independent of product-specific implementation details.

\section{Experimental Evaluation}

We evaluate the hierarchical architecture against a flat-registry baseline using controlled benchmarks that isolate routing architecture while holding the LLM, decision protocol, and scoring methodology constant.

\subsection{Evaluation Goals and Scope}

The evaluation addresses three questions: (1) does hierarchical lazy discovery bound per-step prompt and schema exposure as $N$ grows? (2) under multi-step workflow pressure with noisy chat history, does hierarchy maintain comparable task success with smaller prompts? (3) in production-adjacent UPI flows, how do coverage and latency compare where skill trees are complete versus incomplete?

We do \emph{not} claim universal accuracy superiority for hierarchy, full end-to-end production parity between implementations, or multi-model generalization. Synthetic benchmarks stress catalog scale; real-flow tests measure route-level correctness, not complete business-task completion.

\subsection{Experimental Setup}

\textbf{Architectures.} The flat baseline ($\mathcal{A}_{\text{flat}}$) presents all $N$ tool schemas at every routing step. The hierarchical baseline ($\mathcal{A}_{\text{hierarchical}}$) routes through a fixed skill tree with branching factor $k{=}4$ and depth 2--4, exposing only immediate children at each decision node via localized manifests.

\textbf{Protocol.} Both architectures use the same LLM instance, temperature 0, and plain JSON output validated against a shared Pydantic schema (not provider function-calling), isolating architectural effects from parsing-mode differences.

\textbf{Scenario modes.} We test: \textit{exact} (tool name given verbatim; establishes routing ceiling), \textit{semantic} (overlapping descriptions), \textit{clean\_semantic} (mutually exclusive domains), \textit{agentic} (action + tool + argument extraction), and \textit{workflow} (ordered two-step tasks with four blocks of irrelevant chat history). Synthetic sweeps use $N \in \{16, 32, 64, 128\}$ for routing-accuracy modes (400-char schemas) and $N \in \{16, 64, 128, 256, 512\}$ for the generated scale-up harness (800-char schemas).

\textbf{Metrics.} At the task level we report task success rate and LLM calls per task. At the decision level we report parse, validation, wrong-action, wrong-choice, and wrong-argument rates. For efficiency we report input tokens per task and \textbf{visible tool schema tokens per LLM call}, the tokens attributable to tool metadata (name, description, JSON schema) included in a single model request at step $t$, i.e., the size of the routing menu visible in $\Omega_t$, not the global registry size $N$.

\subsection{Prompt Scalability and Context Feasibility}

Table~\ref{tab:scaleup} reports the generated scale-up harness (800-char schemas per tool, $k{=}4$, tree depth 4 for hierarchy). Flat routing presents all $N$ schemas in one call; hierarchy uses four sequential routing calls at depth 4. Flat input and visible-schema tokens grow approximately linearly with $N$; hierarchical per-step schema exposure remains bounded (138--1,111 tokens per call in this configuration).

\begin{table*}[t]
\centering
\caption{Scale-up routing under increasing catalog size $N$ (generated catalogs, 800-char schemas, hierarchical depth 4).}
\label{tab:scaleup}
\footnotesize
\begin{tabular}{@{}lcccccc@{}}
\toprule
\textbf{Arch.} & $N$ & \textbf{Success} & \textbf{Calls} & \textbf{Input tok.} & \textbf{Schema tok./call} & \textbf{LLM ms} \\
\midrule
\multicolumn{7}{l}{\textit{Flat (all tools visible per call)}} \\
flat & 16 & 1.0 & 1.0 & 3,650 & 2,223 & 702 \\
flat & 64 & 1.0 & 1.0 & 14,162 & 8,895 & 1,511 \\
flat & 128 & 1.0 & 1.0 & 28,178 & 17,791 & 2,584 \\
flat & 256 & 0.0 & 1.0 & 49,787 & 35,583 & --(context limit) \\
flat & 512 & 0.0 & 1.0 & 99,451 & 71,167 & --(context limit) \\
\midrule
\multicolumn{7}{l}{\textit{Hierarchical (lazy children only)}} \\
hierarchical & 16 & 1.0 & 4.0 & 1,332 & 138 & 2,067 \\
hierarchical & 64 & 1.0 & 4.0 & 1,528 & 138 & 2,051 \\
hierarchical & 128 & 1.0 & 4.0 & 1,733 & 277 & 2,087 \\
hierarchical & 256 & 1.0 & 4.0 & 2,147 & 555 & 2,159 \\
hierarchical & 512 & 1.0 & 4.0 & 2,875 & 1,111 & 2,116 \\
\bottomrule
\end{tabular}
\end{table*}

At $N{=}256$ and $N{=}512$, flat routing incurs 100\% parse/validate failure (context limit exceeded), while hierarchy remains feasible with under 3k input tokens per task. At $N{=}128$, flat uses ${\sim}28k$ input tokens versus ${\sim}1.7k$ for hierarchy (${\sim}16\times$ reduction) while both achieve 100\% routing success in this controlled task. A complementary \textit{agentic} stress at $N{=}256$ with 800-char schemas (Table~\ref{tab:n256agentic}) shows flat prompts reaching ${\sim}278k$ characters with 0\% success, versus hierarchical ${\sim}8.9k$ characters per decision with 70\% task success, thus confirming feasibility under realistic schema pressure beyond the generated scale-up harness alone.

\subsection{Workflow Pressure and Routing Errors}

Table~\ref{tab:workflow} reports full \textit{workflow}-mode sweeps with noisy chat history (30 trials per $N$, 800-char schemas, four history-repeat blocks, two ordered tool calls per scenario).

\begin{table*}[t]
\centering
\caption{Workflow routing with noisy chat history ($t{=}30$ trials, all tested $N$).}
\label{tab:workflow}
\footnotesize
\begin{tabular}{@{}lccccccc@{}}
\toprule
\textbf{Arch.} & $N$ & \textbf{Success} & \textbf{Prompt chars} & \textbf{Wrong act.} & \textbf{Wrong choice} & \textbf{Wrong args} & \textbf{Schema tok./call} \\
\midrule
flat & 16 & 0.74 & 18,515 & 0.12 & 0.12 & 0.12 & 2,238 \\
flat & 32 & 0.73 & 35,923 & 0.13 & 0.13 & 0.13 & 4,550 \\
flat & 64 & 0.73 & 70,727 & 0.13 & 0.13 & 0.13 & 9,176 \\
flat & 128 & 0.73 & 140,335 & 0.13 & 0.13 & 0.13 & 18,428 \\
hierarchical & 16 & 0.73 & 1,809 & 0 & 0 & 0.03 & 76 \\
hierarchical & 32 & 0.73 & 2,308 & 0 & 0 & 0.03 & 127 \\
hierarchical & 64 & 0.73 & 3,306 & 0 & 0 & 0.03 & 229 \\
hierarchical & 128 & 0.73 & 5,302 & 0 & 0 & 0.03 & 433 \\
\bottomrule
\end{tabular}
\end{table*}

At $N{\geq}32$ within context limits, both architectures converge to ${\sim}0.73$ task success, but hierarchical prompts remain an order of magnitude smaller (e.g., 5.3k vs.\ 140k characters at $N{=}128$, a ${\sim}26\times$ reduction). Hierarchical runs exhibit near-zero wrong-action and wrong-choice rates; residual failures are primarily wrong arguments (${\sim}0.027$). Workflow stress runs use more sequential LLM calls for hierarchy (${\sim}10$ calls/task vs.\ ${\sim}2$ for flat at $N{=}128$) but keep total input tokens per task lower (10.9k vs.\ 56.4k in the efficiency harness).

\subsection{Agentic Routing and High-Schema Pressure}

Table~\ref{tab:agentic} summarizes \textit{agentic} mode (action + tool + argument extraction, corrected schema, 400-char payloads, 10 trials per $N$). Both architectures achieve ${\sim}0.7$--$0.8$ task success at $N{\leq}128$, but flat wrong-action and wrong-argument rates are roughly $4\times$ higher than hierarchy (0.2 vs.\ 0.048 per decision).

\begin{table}[t]
\centering
\caption{Agentic routing across tool count $N$ (400-char schemas).}
\label{tab:agentic}
\footnotesize
\begin{tabular}{@{}lccccc@{}}
\toprule
\textbf{Arch.} & $N$ & \textbf{Success} & \textbf{Prompt chars} & \textbf{Wrong act.} & \textbf{Wrong args} \\
\midrule
flat & 16 & 0.90 & 10,133 & 0.0 & 0.10 \\
flat & 32 & 0.70 & 21,135 & 0.2 & 0.30 \\
flat & 64 & 0.70 & 43,139 & 0.2 & 0.30 \\
flat & 128 & 0.50 & 87,147 & 0.3 & 0.40 \\
hier. & 16 & 1.00 & 1,117 & 0.0 & 0.0 \\
hier. & 32 & 0.70 & 1,476 & 0.05 & 0.05 \\
hier. & 64 & 0.70 & 2,193 & 0.05 & 0.05 \\
hier. & 128 & 0.70 & 3,628 & 0.05 & 0.05 \\
\bottomrule
\end{tabular}
\end{table}

\begin{table}[t]
\centering
\caption{Agentic routing at $N{=}256$ with 800-char schemas (context-limit stress).}
\label{tab:n256agentic}
\footnotesize
\begin{tabular}{@{}lccccc@{}}
\toprule
\textbf{Arc.} & \textbf{Suc.} & \textbf{PromptChars} & \textbf{ParseFail} & \textbf{WrongAct} & \textbf{WrongArgs} \\
\midrule
flat & 0.0 & 277,659 & 1.0 & 0.0 & 0.0 \\
hier. & 0.7 & 8,867 & 0.0 & 0.05 & 0.05 \\
\bottomrule
\end{tabular}
\end{table}

\subsection{Semantic and Clean-Semantic Routing}

We evaluate routing under semantic pressure using two diagnostic modes ($k{=}4$, 30 trials per $N$, 400-char schemas). \textit{Semantic} mode uses overlapping tool descriptions so multiple tools might sound alike or use the same keywords (Ex. 'payment,' 'fraud,' 'FAQ,' and 'mandate')and the user request doesn’t explicitly name which tool it needs. While \textit{clean\_semantic} mode uses mutually exclusive domains (cards, loans, investments, insurance) and prompts list the names of the options apart from their descriptions to make the choices less ambiguous. These tests isolate \textbf{hierarchy design quality}, not general architecture superiority: flat sees full tool-level evidence in one prompt while hierarchy makes lossy top-level choices first.

\textbf{Semantic (overlapping domains).} Table~\ref{tab:semantic} shows flat routing at 100\% success for $N{=}16$--$128$ as prompts grow to 88.2k characters, which is expected when the full catalog fits in context and requests are clean synthetic retrieval tasks. Hierarchy scores 0.83/0.83/0.67/0.73 across $N$, with wrong-choice rates of 0.06--0.14 and zero parse or validate failures. All hierarchical failures occur at the \textbf{first routing depth}: the dominant pattern is \texttt{mandate} expected but \texttt{payment} selected (e.g., ``autopay subscriptions''), which removes the correct branch before leaf tools are visible. This supports a design warning about overlapping domain boundaries, not a claim that hierarchy is inherently inferior.

\begin{table*}[t]
\centering
\caption{Semantic overlapping-intent routing ($k{=}4$, 400-char schemas, 30 trials per $N$).}
\label{tab:semantic}
\footnotesize
\begin{tabular}{@{}lccccccc@{}}
\toprule
\textbf{Arch.} & $N$ & \textbf{Success} & \textbf{Prompt chars} & \textbf{Options} & \textbf{Parse fail} & \textbf{Valid. fail} & \textbf{Wrong choice} \\
\midrule
flat & 16 & 1.0 & 9,507 & 16.0 & 0.0 & 0.0 & 0.0 \\
flat & 32 &  1.0 & 20,770 & 32.0 & 0.0 & 0.0 & 0.0 \\
flat & 64 &  1.0 & 43,259 & 64.0 & 0.0 & 0.0 & 0.0 \\
flat & 128 & 1.0 & 88,243 & 128.0 & 0.0 & 0.0 & 0.0 \\
hierarchical & 16 & 0.83 & 1,236 & 3.1 & 0.0 & 0.0 & 0.06 \\
hierarchical & 32 & 0.83 & 1,814 & 3.4 & 0.0 & 0.0 & 0.06 \\
hierarchical & 64 & 0.67 & 3,049 & 4.7 & 0.0 & 0.0 & 0.14 \\
hierarchical & 128 & 0.73 & 6,084 & 7.3 & 0.0 & 0.0 & 0.11 \\
\bottomrule
\end{tabular}
\end{table*}

\textbf{Clean semantic.} Table~\ref{tab:cleansemantic} shows flat routing still at 100\% success; hierarchy reaches 1.0/0.97/0.87/0.77 at $N{=}16$--$128$ with \textbf{zero wrong-choice errors}. Remaining failures are validate-format errors (0--0.08): the model returns descriptions appended to option names rather than selecting the wrong branch. Prompts remain ${\sim}17\times$ smaller than flat at $N{=}128$ (5.1k vs.\ 86.3k characters), indicating that tree design and prompt discipline matter more than architecture alone.

\begin{table*}[t]
\centering
\caption{Clean semantic routing ($k{=}4$, 400-char schemas, 30 trials per $N$).}
\label{tab:cleansemantic}
\footnotesize
\begin{tabular}{@{}lccccccc@{}}
\toprule
\textbf{Arch.} & $N$ & \textbf{Success} & \textbf{Prompt chars} & \textbf{Options} & \textbf{Parse fail} & \textbf{Valid. fail} & \textbf{Wrong choice} \\
\midrule
flat & 16 & 1.0 & 9,304 & 16.0 & 0.0 & 0.0 & 0.0 \\
flat & 32 & 1.0 & 20,320 & 32.0 & 0.0 & 0.0 & 0.0 \\
flat & 64 & 1.0 & 42,326 & 64.0 & 0.0 & 0.0 & 0.0 \\
flat & 128 & 1.0 & 86,339 & 128.0 & 0.0 & 0.0 & 0.0 \\
hierarchical & 16 & 1.0 & 1,073 & 2.7 & 0.0 & 0.0 & 0.0 \\
hierarchical & 32 & 0.97 & 1,705 & 3.3 & 0.0 & 0.01 & 0.0 \\
hierarchical & 64 & 0.87 & 2,899 & 4.4 & 0.0 & 0.05 & 0.0 \\
hierarchical & 128 & 0.77 & 5,147 & 6.4 & 0.0 & 0.08 & 0.0 \\
\bottomrule
\end{tabular}
\end{table*}

Table~\ref{tab:semanticcompare} summarizes hierarchical success across modes. Clean domain design closes most of the overlap deficit (e.g., $N{=}128$: 0.73$\rightarrow$0.77) by eliminating wrong-branch errors; the residual gap is largely prompt-format hygiene.

\begin{table}[t]
\centering
\caption{Hierarchical task success: semantic vs.\ clean\_semantic ($\Delta$ = clean $-$ semantic).}
\label{tab:semanticcompare}
\footnotesize
\begin{tabular}{@{}ccccc@{}}
\toprule
$N$ & \textit{semantic} & \textit{clean\_semantic} & $\Delta$ & \textbf{Flat (both)} \\
\midrule
16 & 0.83 & 1.00 & +0.17 & 1.0 \\
32 & 0.83 & 0.97 & +0.14 & 1.0 \\
64 & 0.67 & 0.87 & +0.20 & 1.0 \\
128 & 0.73 & 0.77 & +0.03 & 1.0 \\
\bottomrule
\end{tabular}
\end{table}

Production router tests (Table~\ref{tab:realrouter}) echo the overlapping-domain pattern: orchestrator success is 0.925 overall, with autopay-subscription phrasing at 70\% (3/10 misroutes to payment/QA) while clear-intent families reach 100\%.

\begin{table}[t]
\centering
\caption{Production router-only benchmark (real prompts, 10 trials per router).}
\label{tab:realrouter}
\footnotesize
\begin{tabular}{@{}lcc@{}}
\toprule
\textbf{Router} & \textbf{Success} & \textbf{Parse/validate fail} \\
\midrule
orchestrator & 0.925 & 0.0 \\
payment & 0.950 & 0.05 \\
mandate & 1.000 & 0.0 \\
fraud & 0.967 & 0.0 \\
\bottomrule
\end{tabular}
\end{table}

\subsection{Production-Adjacent UPI Flows}

We additionally benchmarked production-adjacent implementations on a 45-query UPI Help suite (mandate, payment, fraud, FAQ, boundary phrasing), 5 trials per query, measuring \textbf{route-level} success. Table~\ref{tab:realflow} reports the \textbf{observed} results.

\begin{table*}[t]
\centering
\caption{Real-flow route-level success by query category (45-query suite, 5 trials each).}
\label{tab:realflow}
\footnotesize
\begin{tabular}{@{}lcccccc@{}}
\toprule
\textbf{Category} & \textbf{System} & \textbf{Success} & \textbf{Calls/q} & \textbf{Tokens/q} & \textbf{LLM ms/q} & \textbf{Wall ms/q} \\
\midrule
mandate\_summary & hierarchical & 1.00 & 3.10 & 5,690 & 1,477 & 1,486 \\
mandate\_summary & flat terminal & 0.70 & 2.50 & 4,606 & 2,511 & 2,946 \\
mandate\_action & hierarchical & 1.00 & 3.40 & 6,452 & 1,821 & 1,832 \\
mandate\_action & flat terminal & 1.00 & 3.00 & 5,602 & 2,967 & 3,100 \\
payment & hierarchical & 1.00 & 5.40 & 9,711 & 2,627 & 2,641 \\
payment & flat terminal & 0.90 & 1.60 & 3,236 & 2,248 & 3,446 \\
fraud & hierarchical & 0.97 & 2.20 & 2870 & 2,130 & 2,150 \\
fraud & flat terminal & 1.00 & 2.00 & 2,870 & 2,130 & 2,158 \\
qa & hierarchical & 1.00 & 2.50 & 4,000 & 1,500 & 1,520 \\
qa & flat terminal & 1.00 & 2.60 & 8,175 & 5,972 & 9,093 \\
boundary & hierarchical & 0.40 & 4.00 & 6,840 & 1,997 & 2,008 \\
boundary & flat terminal & 0.60 & 1.80 & 3,088 & 2,238 & 2,648 \\
\bottomrule
\end{tabular}
\end{table*}

The \textbf{observed} hierarchical skill-tree implementation seems strong and fast with lower LLM latency than flat, with overall success of \textbf{0.93}, comparable to flat (0.87) while retaining faster LLM-time routing on covered paths; boundary phrasing (0.40 hierarchical vs.\ 0.60 flat) remains the main unresolved weakness for both systems. The flat terminal baseline still shows higher wall-clock latency overall due to database and tool runtime overhead, and instability on some mandate detail queries (autopay subscription phrasing). This comparison is not apples-to-apples end-to-end: it reflects current deployment coverage as much as architecture. The UPI Help motivation (Section~8) should be read 
together with these coverage gaps.

\subsection{Discussion}

Choose hierarchical routing when tool catalogs may grow, schemas are rich, multi-step workflows matter, and domain categories can be kept mutually exclusive. Choose flat routing when $N$ is bounded, tool descriptions are distinct, and operational simplicity outweighs scale headroom. Flat routing can be very accurate on clean synthetic retrieval while the full catalog fits in context (Section~9), but becomes risky as $N$, schema size, and history grow. Hierarchy buys context feasibility and bounded decision space; accuracy requires deliberate tree design, non-overlapping top-level domains, and strict option-name output discipline.

\section{Complexity and Performance Analysis}

Empirical evaluation (Section~9) confirms that per-step visible schema exposure is bounded under hierarchy and that flat routing becomes context-infeasible at large $N$ under rich schemas. The following analysis formalizes these observations in terms of prompt size, inference cost, stack memory, and discovery overhead.

\subsection{Prompt Complexity and Token Efficiency}

Let $N = |V|$ be the total number of tools, and let $S_{avg}$ be the average token length of a tool schema. In a standard flat-agent architecture containing $N$ executable tools, the inference engine must embed the metadata and schemas for all $N$ tools into the system prompt during every execution cycle. The prompt complexity per decision step is $\mathcal{O}(N \cdot S_{avg})$. As $N$ grows, this term can dominate the context window and cause context saturation.

In the hierarchical architecture, assume a balanced $k$-ary skill tree, where $k$ is the maximum local branching factor. The depth of the tree is approximately $\log_k N$. At any given decision step, the lazy discovery protocol exposes at most $k$ child schemas. Thus, the prompt complexity per reasoning cycle is bounded by $\mathcal{O}(k \cdot S_{avg})$, which is entirely independent of the global registry size $N$. The total schema-token cost to reach a leaf node is bounded by:

\begin{equation}
C_{\text{total}} = \sum_{d=1}^{\log_k N} k \cdot S_{avg} = \mathcal{O}(k \cdot S_{avg} \cdot \log_k N)
\end{equation}

If $S_{avg}$ is treated as a bounded constant, this simplifies to $\mathcal{O}(k \log_k N)$. This logarithmic path cost allows large capability sets to be handled without exposing the model to the full registry at each decision step, thereby preventing attention bottlenecks and avoiding token exhaustion.

\subsection{Inference Latency and Time Complexity}

The computational cost of transformer-based LLMs scales approximately quadratically with input sequence length because of self-attention. In a flat architecture, the schema portion of the prompt has length $\mathcal{O}(N \cdot S_{avg})$, giving a schema-related attention cost of approximately $\mathcal{O}((N \cdot S_{avg})^2)$ per reasoning step. In the hierarchical architecture, the corresponding per-step schema cost is approximately $\mathcal{O}((k \cdot S_{avg})^2)$.

However, the hierarchy introduces a latency trade-off. Smaller prompts reduce per-step inference cost, but the total number of sequential LLM calls scales proportionally with tree depth. Therefore, optimizing the tree layout (placing high-frequency executables closer to the root while keeping the local branching factor $k$ manageable) is an important design consideration \cite{autoflow, anaflow}.

\subsection{Space Complexity and Stack Memory Bounds}
Traditional ReAct agents \cite{yao2023react} maintain a flat, rolling dialogue history. As the agent explores deep multi-tool workflows, the memory array grows monotonically, yielding a space complexity of $O(A)$ where $A$ is the total number of actions taken.

Our architecture uses a LIFO stack $\Gamma$ coupled with localized output accumulators $\mathcal{O}_{\mathit{acc}}$ residing within each frame. Let $D$ be the maximum depth of the capability tree, and let $B$ bound the retained output footprint per frame (for example, the number or summarized size of child outputs). At any point during execution, the active stack contains at most $D$ frames, yielding an active memory bound of $O(D \cdot B)$. When an orchestrator concludes its localized task and returns its aggregated output, its frame and intermediate scratchpad data can be discarded, preventing unbounded context growth.

Under lazy manifest loading, \emph{active orchestration memory} (loaded schemas, manifests, stack frames, and intermediate outputs along the current path) scales with local branching and depth rather than global registry size $N$, whereas a flat agent that eagerly materializes the full tool registry incurs $\mathcal{O}(N)$ active metadata when all schemas are held in memory. The stack-frame overhead is negligible relative to large tool schemas at enterprise scale.

\subsection{Discovery Overhead}
Lazy discovery introduces a localized parsing step (Algorithm 2) to load the child manifest at $\rho$ when entering a new decision node. This adds a disk I/O overhead bounded by $O(1)$ per traversal step, which is typically small compared to the reduction in LLM inference cost from omitting irrelevant tools. Furthermore, simple memory-caching of the discovered metadata amortizes this disk access cost to near zero on subsequent traversals.

\section{System Constraints and Limitations}

Despite the structural predictability and security enhancements introduced by the framework, the reliance on language models and localized memory arrays presents specific theoretical constraints that warrant critical analysis.

\subsection{Localized Token Accumulation Limits}
While capability fencing aggressively reduces the size of the immediate tool schema provided to the LLM, the accumulation of execution history presents an inherent token scaling constraint. By replacing global output accumulation with localized, per-frame accumulators ($\mathcal{O}_{\mathit{acc}}$), the architecture successfully bounds the rolling context window vertically; a parent node only processes the outputs of its immediate children, preventing deep-tree traversals from saturating the orchestrator's token limits. However, the system assumes that the compiled sequence of localized child responses will remain within the context window of the immediate parent. For exceptionally wide fan-out workflows, where a single decision-making node delegates to dozens of executable sub-skills, token exhaustion remains a vulnerability. Future iterations addressing massive fan-outs may require the integration of modular episodic summarization techniques \cite{park2023generative}.

\subsection{Probabilistic Flow Control and State Thrashing}
A critical constraint of the current implementation is reliance on the LLM to signal whether a delegated task is complete. Because this completion flag is produced by a probabilistic model, it introduces a semantic failure mode into an otherwise deterministic control loop. While the stack-pending invariant ensures that premature or incorrect returns simply hand control back to the parent (enabling retries), repeated misclassification of completion can cause \emph{local oscillation} within a branch (``thrashing''), potentially leading to long or non-terminating loops. A stronger solution is to add explicit validators or constraint checkers \cite{li2024formalllm} that verify completion conditions programmatically before allowing a stack transition.

\subsection{Benchmark and Evaluation Limitations}
The experimental evaluation (Section~9) uses a single LLM and inference-server configuration; results may not transfer across models. Synthetic catalogs stress scale ($N$, schema size) more than production messiness (overlapping policies, boundary phrasing, stateful corrections). Route-level success does not imply end-to-end business-task completion. Real-flow comparisons between implementations reflect different domain coverage (e.g., missing fraud/QA routes in the hierarchical tree) and are not a controlled architectural isolate. Hierarchy accuracy is sensitive to overlapping top-level domains and prompt-format discipline, not inherently to depth or lazy loading, thus poorly designed trees can underperform flat routing despite smaller prompts.

\section{Conclusion}

This paper presented the formalization and implementation of a hierarchical, skill-based architecture for orchestrating arbitrary-depth multi-agent workflows. By replacing monolithic tool registries with a localized, manifest-driven capability tree, the framework reduces decision-space explosion and supports lazy capability loading. A uniform, stack-based execution loop aligns the framework with a context-free control model (Pushdown Automata), enabling deterministic workflow resumption and explicit state tracking.

Separating execution logic from capability declarations allows the system to scale via declarative manifest updates without altering the orchestration engine. Capability fencing and localized execution boundaries also reduce the risk of indirect prompt injection, supporting governance requirements in high-trust enterprise environments. Controlled benchmarks show that hierarchical routing bounds visible schema tokens per LLM call and remains context-feasible at $N{=}512$ with rich schemas, where flat routing fails; under workflow pressure, hierarchy can match flat task success while reducing prompt size by an order of magnitude. Semantic tests further show that hierarchy is sensitive to overlapping domain design but does not collapse at moderate $N$ when trials are sufficient; clean hierarchies eliminate wrong-branch errors, leaving mostly fixable format-validation failures. The UPI Help deployment context illustrates how these requirements arise in real payment-support systems. Overall, the architecture provides a practical blueprint for moving LLM-based agents from prototypes to auditable and scalable production systems.

\section{Contributors}

\noindent\textbf{Leadership:} Prashant Devadiga

\medskip
\noindent\textbf{Agentic Architecture:} Abhishek, Alok Singh, Asit Desai, Gaurang Dahad, Harshit Bhushan, Prakhar Gupta

\medskip
\noindent\textbf{Engineering:} Adithya Mishra, Amisha Sinha, Mandati Pramod Reddy, Rupesh Patil, Siddhi Behere

\medskip
\noindent\textbf{Evaluation:} Abhishek, Gaurang Dahad, Harshit Bhushan


\begin{thebibliography}{33}

\bibitem{vaswani2017attention}
A.~Vaswani \emph{et al.}, ``Attention Is All You Need,'' in \emph{Proc. Advances in Neural Information Processing Systems (NeurIPS)}, vol.~30, 2017.

\bibitem{schick2023toolformer}
T.~Schick \emph{et al.}, ``Toolformer: Language Models Can Teach Themselves to Use Tools,'' \emph{arXiv preprint arXiv:2302.04761}, 2023.

\bibitem{liu2023agentbench}
X.~Liu \emph{et al.}, ``AgentBench: Evaluating LLMs as Agents,'' in \emph{Proc. International Conference on Learning Representations (ICLR)}, 2024.

\bibitem{qin2023toolllm}
Y.~Qin \emph{et al.}, ``ToolLLM: Facilitating Large Language Models to Master 16000+ Real-world APIs,'' in \emph{Proc. International Conference on Learning Representations (ICLR)}, 2024.

\bibitem{du2024anytool}
Y.~Du, F.~Wei, and H.~Zhang, ``AnyTool: Self-Reflective, Hierarchical Agents for Large-Scale API Calls,'' in \emph{Proc. International Conference on Machine Learning (ICML)}, 2024.

\bibitem{xu2024toolbench}
Q.~Xu \emph{et al.}, ``On the Tool Manipulation Capability of Open-source Large Language Models,'' \emph{arXiv preprint arXiv:2305.16504}, 2023.

\bibitem{wu2023autogen}
Q.~Wu \emph{et al.}, ``AutoGen: Enabling Next-Gen LLM Applications via Multi-Agent Conversation Framework,'' \emph{arXiv preprint arXiv:2308.08155}, 2023.

\bibitem{hong2023metagpt}
S.~Hong \emph{et al.}, ``MetaGPT: Meta Programming for A Multi-Agent Collaborative Framework,'' in \emph{Proc. International Conference on Learning Representations (ICLR)}, 2024.

\bibitem{qian2023chatdev}
C.~Qian \emph{et al.}, ``Communicative Agents for Software Development,'' in \emph{Proc. Annual Meeting of the Association for Computational Linguistics (ACL)}, 2024.

\bibitem{packer2023memgpt}
C.~Packer \emph{et al.}, ``MemGPT: Towards LLMs as Operating Systems,'' \emph{arXiv preprint arXiv:2310.08560}, 2023.

\bibitem{npciupihelp}
National Payments Corporation of India, ``UPI Help,'' 2025. [Online]. Available: \url{https://upihelp.npci.org.in}. [Accessed: May~12, 2026].

\bibitem{sipser2012introduction}
M.~Sipser, \emph{Introduction to the Theory of Computation}, 3rd~ed. Cengage Learning, 2012.

\bibitem{liang2023taskmatrix}
Y.~Liang \emph{et al.}, ``TaskMatrix.AI: Completing Tasks by Connecting Foundation Models with Millions of APIs,'' \emph{arXiv preprint arXiv:2303.16434}, 2023.

\bibitem{wang2023voyager}
G.~Wang \emph{et al.}, ``Voyager: An Open-Ended Embodied Agent with Large Language Models,'' \emph{arXiv preprint arXiv:2305.16291}, 2023.

\bibitem{kim2024llmcompiler}
S.~Kim \emph{et al.}, ``An LLM Compiler for Parallel Function Calling,'' in \emph{Proc. International Conference on Machine Learning (ICML)}, 2024.

\bibitem{shen2023hugginggpt}
Y.~Shen, K.~Song, X.~Tan, D.~Li, W.~Lu, and Y.~Zhuang, ``HuggingGPT: Solving AI Tasks with ChatGPT and Its Friends in Hugging Face,'' in \emph{Proc. Advances in Neural Information Processing Systems (NeurIPS)}, vol.~36, 2023.

\bibitem{zhang2023agentverse}
W.~Chen \emph{et al.}, ``AgentVerse: Facilitating Multi-Agent Collaboration and Exploring Emergent Behaviors,'' in \emph{Proc. International Conference on Learning Representations (ICLR)}, 2024.

\bibitem{anthropic2025mcp}
Anthropic, ``Model Context Protocol Specification,'' 2025. [Online]. Available: \url{https://modelcontextprotocol.io}. [Accessed: Mar.~13, 2026].

\bibitem{graefe1993query}
G.~Graefe, ``Query Evaluation Techniques for Large Databases,'' \emph{ACM Computing Surveys (CSUR)}, vol.~25, no.~2, pp.~73--170, 1993.

\bibitem{chomsky1956three}
N.~Chomsky, ``Three Models for the Description of Language,'' \emph{IRE Transactions on Information Theory}, vol.~2, no.~3, pp.~113--124, 1956.

\bibitem{yao2023react}
S.~Yao \emph{et al.}, ``ReAct: Synergizing Reasoning and Acting in Language Models,'' in \emph{Proc. International Conference on Learning Representations (ICLR)}, 2023.

\bibitem{wei2022chain}
J.~Wei \emph{et al.}, ``Chain-of-Thought Prompting Elicits Reasoning in Large Language Models,'' in \emph{Proc. Advances in Neural Information Processing Systems (NeurIPS)}, vol.~35, 2022.

\bibitem{park2023generative}
J.~S.~Park \emph{et al.}, ``Generative Agents: Interactive Simulacra of Human Behavior,'' in \emph{Proc. 36th Annual ACM Symposium on User Interface Software and Technology (UIST)}, 2023.

\bibitem{li2024formalllm}
Z.~Li \emph{et al.}, ``Formal-LLM: Integrating Formal Language and Natural Language for Controllable LLM-based Agents,'' \emph{arXiv preprint arXiv:2402.00798}, 2024.

\bibitem{yao2023tot}
S.~Yao \emph{et al.}, ``Tree of Thoughts: Deliberate Problem Solving with Large Language Models,'' in \emph{Proc. Advances in Neural Information Processing Systems (NeurIPS)}, 2023.

\bibitem{openapi2021}
OpenAPI Initiative, ``OpenAPI Specification v3.1.0,'' 2021. [Online]. Available: \url{https://spec.openapis.org/oas/v3.1.0}. [Accessed: Mar.~13, 2026].

\bibitem{pezoa2016foundations}
F.~Pezoa \emph{et al.}, ``Foundations of JSON Schema,'' in \emph{Proc. 25th International Conference on World Wide Web (WWW)}, 2016.

\bibitem{saltzer1975protection}
J.~H.~Saltzer and M.~D.~Schroeder, ``The Protection of Information in Computer Systems,'' \emph{Proceedings of the IEEE}, vol.~63, no.~9, pp.~1278--1308, 1975.

\bibitem{perez2022ignore}
F.~Perez and I.~Ribeiro, ``Ignore Previous Prompt: Attack Techniques for Language Models,'' \emph{arXiv preprint arXiv:2211.09527}, 2022.

\bibitem{liu2023promptinjection}
Y.~Liu \emph{et al.}, ``Prompt Injection Attack Against LLM-Integrated Applications,'' \emph{arXiv preprint arXiv:2306.05499}, 2023.

\bibitem{openai2023gpt4}
OpenAI, ``GPT-4 Technical Report,'' \emph{arXiv preprint arXiv:2303.08774}, 2023.

\bibitem{autoflow}
J.~Zhang \emph{et al.}, ``AFlow: Automating Agentic Workflow Generation,'' \emph{arXiv preprint arXiv:2410.10762}, 2024.

\bibitem{anaflow}
H.~Zhou \emph{et al.}, ``AnaFlow: Agentic LLM-based Workflow for Reasoning-Driven Explainable Analog Circuit Sizing,'' \emph{arXiv preprint arXiv:2502.02533}, 2025.

\end{thebibliography}
\end{document}